\documentclass{article}
\usepackage{spconf,amsmath,graphicx,amssymb,cite,hyperref,multirow,booktabs,array}
\usepackage{subcaption}
\usepackage{caption}
\usepackage{enumitem}
\setlist{nosep, leftmargin=14pt}

\usepackage{mwe} % to get dummy images

\title{Fast, Unsupervised Framework for Registration Quality Assessment of Multi-stain Histological Whole Slide Pairs}
\name{\centering Shikha Dubey$^{1}$, Patricia Raciti$^{2}$, Kristopher Standish$^{1}$, Albert Juan Ramon$^{1}$, Erik Ames Burlingame$^{\star 1}$  \thanks{$^{\star}$ Corresponding author, eburlin1@its.jnj.com}}
\address{$^{1}$Johnson \& Johnson, Data Science and Digital Health, Computer Vision, NJ, USA \\
$^{2}$Johnson \& Johnson, Oncology Translational Research, PA, USA}

\begin{document}
%\ninept
%
\maketitle
\begingroup
\renewcommand\thefootnote{}
\footnotetext{© 2026 IEEE. Personal use of this material is permitted. Permission from IEEE must be obtained for all other uses, in any current or future media, including reprinting/republishing this material for advertising or promotional purposes, creating new collective works, for resale or redistribution to servers or lists, or reuse of any copyrighted component of this work in other works.}
\addtocounter{footnote}{-1}
\endgroup
\begin{abstract}
High-fidelity registration of histopathological whole slide images (WSIs), such as  hematoxylin \& eosin (H\&E) and immunohistochemistry (IHC), is vital for integrated molecular analysis but challenging to evaluate without ground-truth (GT) annotations. Existing WSI-level assessments—using annotated landmarks or intensity-based similarity metrics—are often time-consuming, unreliable, and computationally intensive, limiting large-scale applicability. This study proposes a fast, unsupervised framework that jointly employs down-sampled tissue masks- and deformations-based metrics for registration quality assessment (RQA) of registered H\&E and IHC WSI pairs. The masks-based metrics measure global structural correspondence, while the deformations-based metrics evaluate local smoothness, continuity, and transformation realism. Validation across multiple IHC markers and multi-expert assessments demonstrate a strong correlation between automated metrics and human evaluations. In the absence of GT, this framework offers reliable, real-time RQA with high fidelity and minimal computational resources, making it suitable for large-scale quality control in digital pathology.\end{abstract}
\begin{keywords}
Image Registration, Quality Control, Ground-truth-free Evaluation, Deformation-fields Analysis
\end{keywords}
\section{Introduction}
\label{intro}
Reliable registration across histological modalities—such as hematoxylin and eosin (H\&E) \cite{GeoMetrics3}, immunohistochemistry (IHC) \cite{anhir,ACROBAT}, autofluorescence, and immunofluorescence \cite{valis}—is essential for integrating structural, molecular, and spatial information in digital pathology.  Accurate alignment enables downstream analyses like cell-type mapping, cross-marker quantification, and virtual multiplexing \cite{vims}. However, evaluating registration quality (RQ) at the whole-slide image (WSI) scale remains computationally expensive: each slide contains billions of pixels, making pixel-wise validation both memory- and time-intensive \cite{elastix,GeoMetrics3,anhir,ACROBAT,valis,deephistreg}. Despite significant progress in registration algorithms \cite{valis,ants,voxelmorph,elastix,deephistreg}, assessing histological RQ remains challenging. Most frameworks \cite{GeoMetrics,GeoMetrics2Path,GeoMetrics3} rely on intensity-based metrics such as mutual information (MI), cross-correlation, mean square error (MSE), and structural similarity (SSIM) that perform well on mono-stain datasets \cite{GeoMetrics3} but fail across stains due to color and texture heterogeneity. Studies like ANTs \cite{ants} focus on 3D MRI and CT, using relative thickness measures, while Avants et al.\cite{deform} assessed registration quality via volume correlation with ground truth (GT). These methods are not directly applicable to 2D histopathology. \\
Landmark-based \cite{ACROBAT,anhir,GeoMetrics2Path} and manually-masked \cite{voxelmorph,PATIL2023,elastix,GeoMetrics2Path} evaluations are among the most widely used, including in recent frameworks such as DeepHistReg \cite{deephistreg} and VALIS \cite{valis}, where RQ is measured via target registration error (TRE) from manually annotated correspondences, which is infeasible for large-scale WSI analysis. Region-overlap metrics (DICE, IoU) in studies \cite{voxelmorph,elastix,PATIL2023} quantify geometric similarity but rely on expert-annotated masks. Hoque et al. \cite{GeoMetrics2Path} used both feature matching and intensity metrics for assessment. However, all these methods ignore deformations plausibility—a critical factor for preserving tissue morphology. \\
\begin{figure}[tb]
\begin{minipage}[b]{0.99\linewidth}
  \centering
  \centerline{\includegraphics[clip, trim=0.0 4.45cm 0.00cm 0.00cm, width=8.1cm]{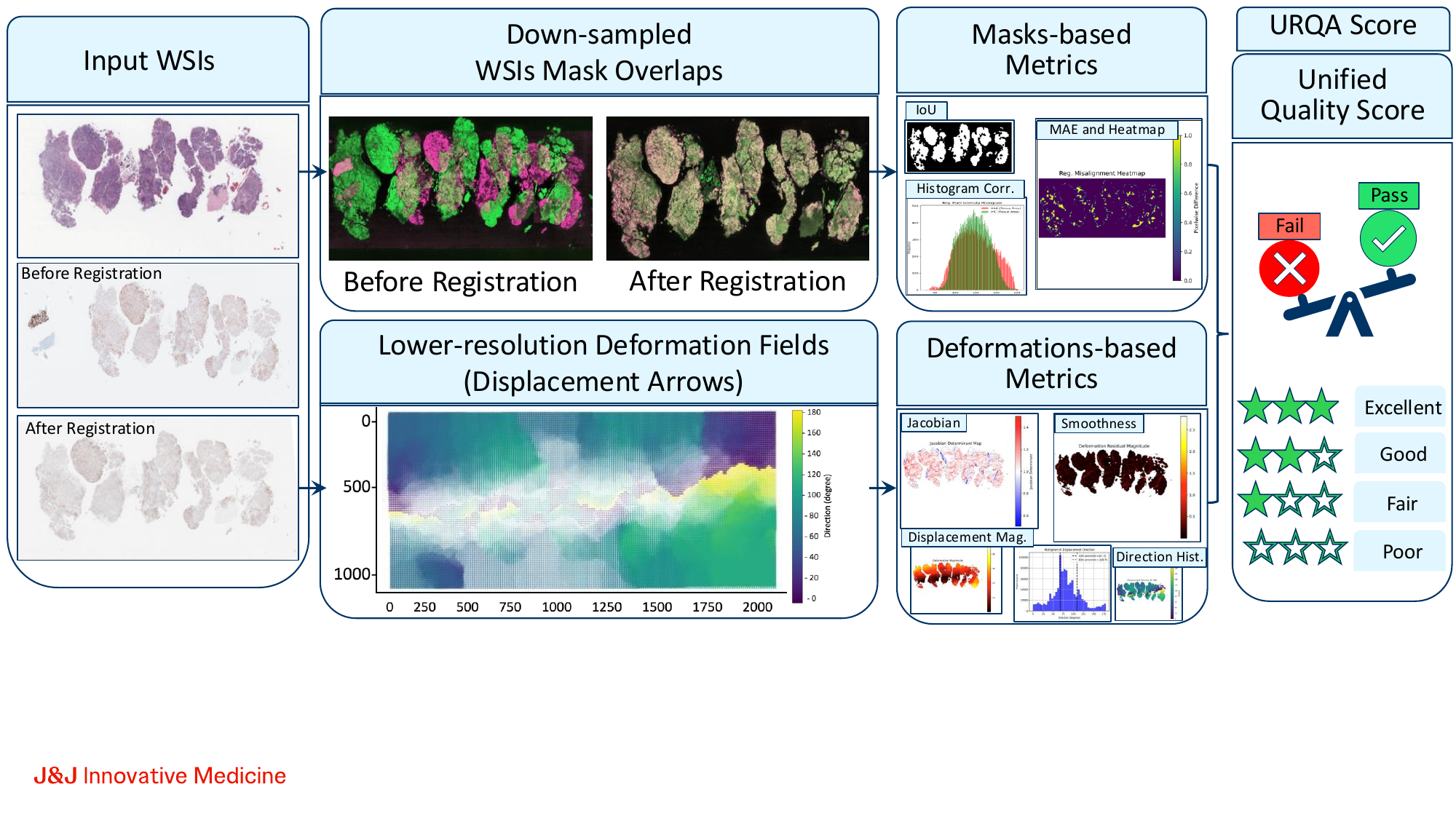}}
  \label{mainFig}
\end{minipage}
\vspace{-0.3cm}
\caption{\small URQA Framework. Mag.: magnitude, Hist: histogram. \normalsize}
\label{mainFig}
\vspace{-0.5cm}
\end{figure}
\hspace{-0.25cm}Deformation-fields analysis has been explored in MRI data \cite{ASYLUM}. Similarly, Cagni et al.\cite{eval1} correlated deformation magnitude with TRE in multimodal CT–CBCT data, but such evaluation has not yet been extended to histological slides. Recent frameworks \cite{voxelmorph,deephistreg,valis} generate deformation fields during registration but use them solely for transformation, not for evaluating RQ.
To our knowledge, no existing method quantifies deformation fields or jointly assesses deformations plausibility alongside geometric alignment metrics in histology, where tissue-level structural integrity is vital.\\
To address these limitations, this work proposes a GT-free registration quality assessment (RQA) framework that jointly measures global tissue alignment and local deformations regularity using lightweight representations which enables rapid, unsupervised evaluation within seconds per slide—bridging the gap between computational efficiency, deformations realism, and biological interpretability in multi-stain RQA. \\
\textbf{Main Contributions:}
\begin{enumerate}
    \item A novel, unsupervised registration quality assessment (URQA) framework for H\&E–IHC WSIs.
    \item Integration of masks-based geometric and deformations-based regularity metrics into a unified score.
    \item Validation through multi-expert assessments, demonstrating efficiency, interpretability, and scalability on multiple IHCs.
\end{enumerate}
\section{Methodology}
\label{method}
The proposed Unsupervised Registration Quality Assessment (URQA) framework, shown in Fig.\ref{mainFig}, evaluates RQ by jointly analyzing tissue geometry—using tissue masks—and deformation-fields regularity between a fixed H\&E and a registered moving IHC image, generated by the VALIS \cite{valis} model, which was ranked as the best among open-source models in ACROBAT 2022 \cite{ACROBAT}. \textbf{Problem Formulation:} $I_f$ and $I_m$ denotes the fixed and moving WSIs, respectively. A registration algorithm estimates a deformation field $\phi$, that warps $I_m$ into the coordinate space of $I_f$. $I_r(x) = I_m(\phi(x))$, where $I_r$ is the registered moving image. This work aims to estimate an URQA score $Q$: $Q(\phi, I_f, I_r)$,without GT, based on the following RQA modules.
\vspace{-0.3cm} \subsection{Masks-based RQA (MRQA) Module} \label{maskthresh}
\begin{figure}[tb]
\begin{minipage}[b]{1.0\linewidth}
  \centering
  \centerline{\includegraphics[clip, trim=0.0 4.6cm 7.0cm 0.00cm, width=6.5cm]{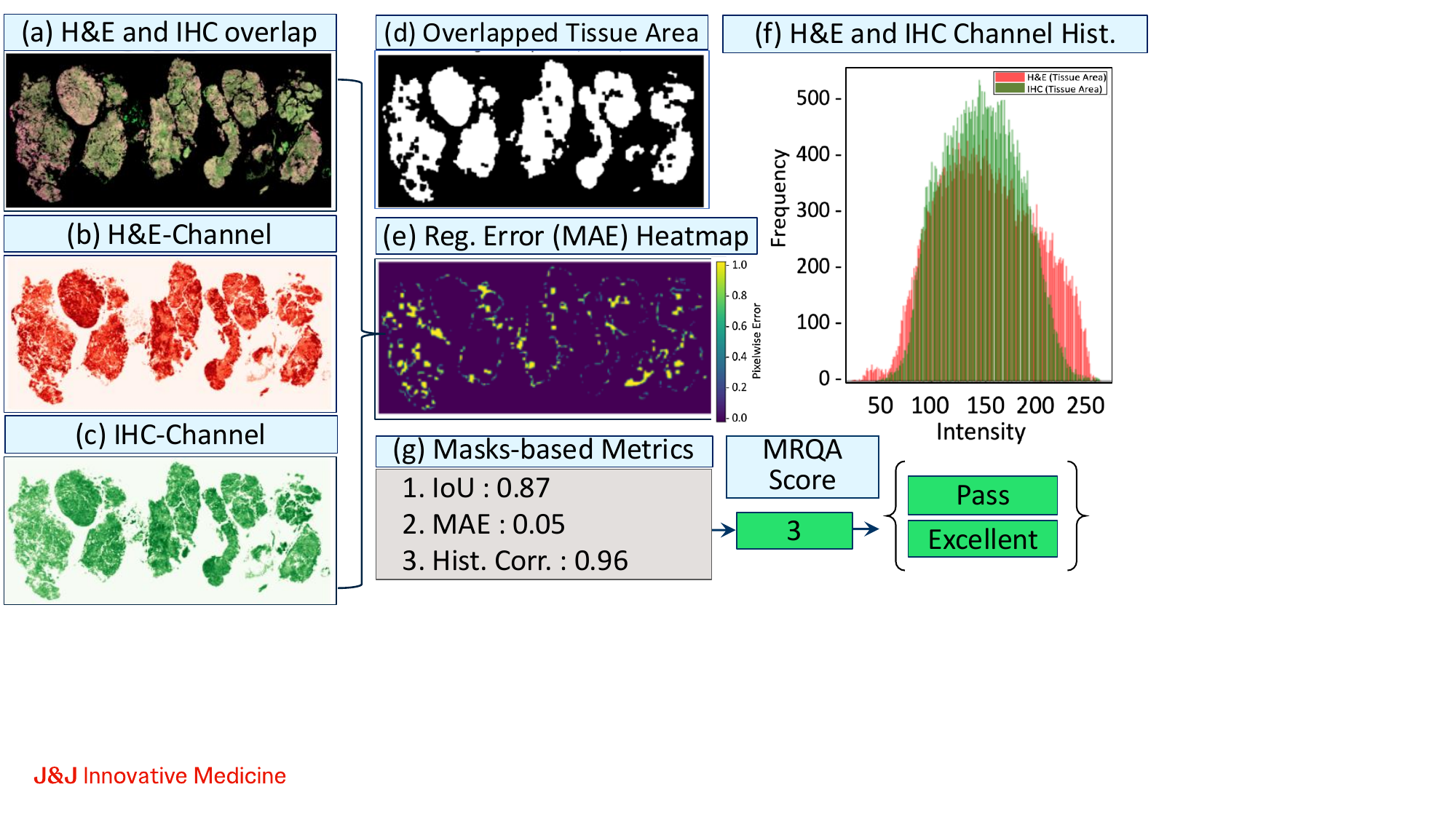}}
\vspace{-0.25cm}
  \centerline{\small (A) Masks-based RQA (MRQA) Module \normalsize}\medskip
 \label{MasksURQA}
 \vspace{-0.2cm}
\end{minipage}
\begin{minipage}[b]{1.0\linewidth}
  \centering
  \centerline{\includegraphics[clip, trim=0.0 6.0cm 0.8cm 0.00cm,width=7.5cm]{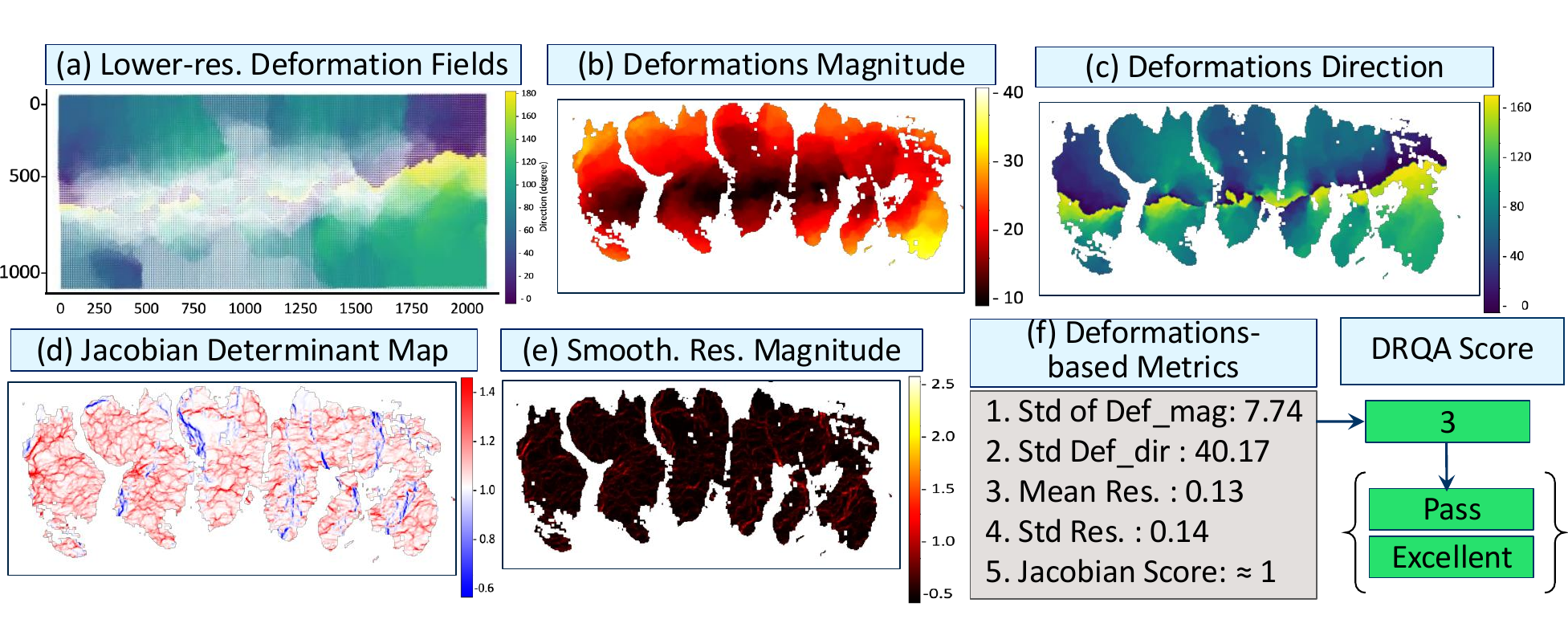}}
    \vspace{-0.2cm}
  \centerline{\small (B) Deformations-based RQA (DRQA) Module \normalsize }\medskip
   \label{DeformURQA}
\end{minipage}
 \vspace{-0.6cm}
\caption{\small (A) and (B) show both modules of URQA. Reg.: Registration; Hist. Corr.: Histogram Correlation; Lower-res.: Lower-resolution; Smooth. Residual: Smoothness residual. \normalsize}
\label{fig2}
\vspace{-0.1cm}
\end{figure}

The MRQA module is shown in Fig.\ref{fig2}(A). Both $I_f$ and $I_r$ are converted into binary tissue masks $M_f$ and $M_r$ using Otsu thresholding \cite{otsu} and morphological operations to isolate tissue regions (Fig.\ref{fig2}(A.a–A.d)). To ensure computational efficiency, both WSIs are down-sampled to 
$min(\text{lowest resolution in WSI pyramid},512)$ before mask generation. This masks-based representation could provide a stain-invariant abstraction that eliminates background bias while preserving overall tissue morphology. The following three metrics are combined to quantify the geometric alignment between $M_f$ and $M_r$. \\
\textbf{Intersection-over-Union (IoU):}$    \text{  IoU} = \frac{|M_f \cap M_r|}{|M_f \cup M_r|}$.\\
\textbf{Mean Absolute Error (MAE):} MAE between tissue masks (Fig.\ref{fig2}(A.e)); lower values indicate better alignment.\\ \vspace{-0.4cm}
\\
\textbf{Histogram Correlation (HC):}
To capture the global distributional similarity between tissue structures, three histogram-based measures are calculated (Fig.\ref{fig2}(A.f)):
\begin{itemize}
    \item \textbf{Pearson correlation coefficient:} \small \vspace{-0.01cm}
    \[
    \text{HC}_{\text{corr}} = 
    \frac{\sum_i (h_f(i) - \bar{h}_f)(h_r(i) - \bar{h}_r)}
    {\sqrt{\sum_i (h_f(i) - \bar{h}_f)^2 \sum_i (h_r(i) - \bar{h}_r)^2}}.
    \] \normalsize \vspace{-0.3cm}
    \item \textbf{Histogram overlap:} \small 
   $
    \text{HC}_{\text{overlap}} = \sum_i \min(h_f(i), h_r(i)).
    $ \normalsize 
    \item \textbf{Cosine similarity:} \small 
    $
    \text{HC}_{\text{cos}} = 
    \frac{h_f \cdot h_r}{\|h_f\|_2 \|h_r\|_2}.
    $\vspace{-0.0cm} \normalsize
\end{itemize}
where $h_f$ and $h_r$ are normalized tissue-area histograms of the down-sampled grayscale images $I_f$ and $I_r$ \cite{valis}. The final HC metric is defined as: \small $
    \text{HC} = \max(\text{HC}_{\text{corr}}, \text{HC}_{\text{overlap}}, \text{HC}_{\text{cos}}).
 \vspace{0.2cm}  $\normalsize \\ 
\hspace{-0.01cm} \textbf{MRQA Scoring:}
The overall MRQA score $M_Q$ is discretized into \textit{Pass} (scores 1–3) and \textit{Fail} (score 0). The score 3 indicates the best RQ (Fig.\ref{fig2}(A.g)). The $M_Q$ is defined as: \vspace{-0.02cm}
\footnotesize
\begin{equation}
M_Q = 
\begin{cases}
3, & \text{if } \text{IoU} \geq 0.80, \text{MAE} \leq 0.07,  \text{HC} \geq 0.80,\\
2, & \text{if } \text{IoU} \geq 0.70, \text{MAE} \leq 0.10,  \text{HC} \geq 0.70,\\
1, & \text{if } \text{IoU} \geq 0.64, \text{MAE} \leq 0.11,  \text{HC} \geq 0.64,\\
0, & \text{if } \text{IoU}  < 0.64, \text{MAE} > 0.11,\text{HC}  < 0.64.\\
\end{cases}
\end{equation}
\normalsize
This adaptive, sample-inductive threshold metric enables robust evaluation into \textit{Pass} and \textit{Fail} categories, with scores 3, 2, 1, and 0 corresponding to \textit{Excellent}, \textit{Good}, \textit{Fair}, and \textit{Poor} RQ, in the absence of GT.
\vspace{-0.3cm} \subsection{Deformations-Based RQA (DRQA) Module}\label{deformthresh}
The DRQA module is shown in Fig.\ref{fig2}(B). To assess the physical plausibility and regularity of the estimated deformations, the framework evaluates multiple statistical properties of the deformation field $\phi$ generated by VALIS \cite{valis}, where $\phi(x) = x + u(x)$, and $u(x) = (u_x, u_y)$ denotes the displacement vector at pixel $x$ . In this work, displacement vectors are derived from the lowest resolution in the hierarchical WSI pyramid, as shown in Fig.\ref{fig2}(B.a).  
The evaluation combines the displacement magnitude and direction, Jacobian determinant, and the smoothness residual to quantify deformations realism.\\
\vspace{0.01cm}\textbf{Displacement Magnitude and Direction:}
The displacement magnitude $M(x)$ and directions $\Theta(x)$ are shown in Fig.\ref{fig2}(B.b-c) and computed as:  \vspace{-0.01cm}
\small
\begin{equation}
    M(x) = \sqrt{u_x(x)^2 + u_y(x)^2}, \hspace{0.01cm}
    \Theta(x) = \tan(u_y(x), u_x(x)).
\end{equation} \normalsize
The interquartile range (IQR) is computed as IQR$(M)= Q_{80}(M) - Q_{30}(M)$. The magnitude is considered regular if: \vspace{-0.01cm}
\small
\begin{equation}
    \sigma_M < \text{IQR}(M), \label{7}
\end{equation}
\normalsize
where $\sigma_M$ denotes the standard deviation of $M(x)$. Eq.\ref{7}  indicates that the spread of magnitudes is consistent with the expected interquartile variation. A similar condition is applied to the directional variance $\sigma_\Theta$, based on its IQR: \vspace{-0.07cm}
\small
\begin{equation}
    \sigma_\Theta < \text{IQR}(\Theta),\label{8}
\end{equation}
\normalsize
ensuring angular consistency across the deformation fields.\\
\textbf{Jacobian Determinant Regularity:}
The Jacobian determinant $J(x)$ of the $\phi(x)$ (Fig.\ref{fig2}(B.d)) is computed as: \vspace{-0.01cm} \small
\begin{equation}
    J(x) = \det(\nabla \phi(x)) = (1 + \frac{\partial u_x}{\partial x})(1 + \frac{\partial u_y}{\partial y}) - \frac{\partial u_x}{\partial y} \frac{\partial u_y}{\partial x}.
\end{equation}
\normalsize
Three Jacobian-based criteria are evaluated:  
(i) the mean $\mu_J$ remains near 1,  
(ii) the standard deviation $\sigma_J$ is below 0.25, and  
(iii) the proportion of negative Jacobians (folding) is below 1.5\%.  
A cumulative Jacobian score $(S_J$) is then assigned:  \vspace{-0.02cm} \small
\begin{equation}
    S_J =
    \begin{cases}
    1, & \text{if } \geq 2 \text{ of the 3 conditions are satisfied,} \\ \label{9}
0, & \text{otherwise.}
\end{cases}
\end{equation}  \vspace{-0.02cm} \normalsize\\
\textbf{Smoothness Residual (SR):}
To measure deformations smoothness, the displacement fields $u_x$ and $u_y$ are smoothed using a Gaussian filter $G_\sigma$, as $u_x^{(s)} = G_\sigma * u_x, u_y^{(s)} = G_\sigma * u_y$. The SR ( shown in Fig.\ref{fig2}(B.e)) is computed as:  \vspace{-0.02cm}  \small
\[
r_x = u_x - u_x^{(s)}, \hspace{0.2cm}  r_y = u_y - u_y^{(s)}, \hspace{0.2cm}  
R(x) = \sqrt{r_x^2 + r_y^2}.
\] \normalsize
SR is considered acceptable if both the mean $\mu_R$ and standard deviation $\sigma_R$ of $R(x)$ are smaller than their IQRs:  \vspace{-0.01cm}  \small
\begin{equation}
    \mu_R < \text{IQR}(R), \hspace{0.2cm} \sigma_R < \text{IQR}(R). \label{10}
\end{equation} \vspace{-0.2cm}\\ 
\normalsize \textbf{DRQA Scoring:}
The deformations plausibility score $D_Q$, shown in Fig.\ref{fig2}(B.f), is based on the number of criteria satisfied by the deformation fields. Each criterion contributes one point:
(1) Consistent displacement magnitude Eq.\ref{7},
(2) Stable displacement direction Eq.\ref{8},
(3) Regular Jacobian determinant Eq.\ref{9},
(4) Low mean smoothness residual Eq.\ref{10},
(5) Low residual variation Eq.\ref{10}. The score $D_Q$ is defined as:  \footnotesize
\begin{equation}
D_Q = 
\begin{cases}
3, & \text{if all 5 criteria are satisfied,} \\
2, & \text{if 4 criteria are satisfied,} \\
1, & \text{if 3 criteria are satisfied,} \\
0, & \text{otherwise.}
\end{cases}
\end{equation} \normalsize
This adaptive, sample-inductive threshold method assigns scores 1–3 (\textit{Pass}), corresponding to \textit{Fair}, \textit{Good}, and \textit{Excellent}, indicating smoother, more realistic deformations, while score 0 (\textit{Fail}) reflects \textit{Poor} or non-physical deformations. The $D_Q$ provides a robust, efficient measure of deformation quality, focusing on smoothness, directional coherence, and tissue plausibility. 
\vspace{-0.3cm} \subsection{Unified Quality Score by URQA}
The overall RQ is determined by combining $M_Q$ and $D_Q$ scores. The final scoring $Q$ follows a hierarchical rule:  \vspace{-0.1cm}  \small
\[
Q = \begin{cases}
0, & \text{if } M_Q = 0 \text{ or } D_Q = 0 \quad (\textit{Fail}), \\
\max(M_Q, D_Q), & \text{if } M_Q, D_Q \in \{1, 2, 3\} \quad (\textit{Pass}).
\end{cases} \vspace{-0.1cm} 
\] \normalsize
where scores 1, 2, and 3 correspond to \textit{Fair}, \textit{Good}, and \textit{Excellent}, respectively. This formulation ensures that any failure in tissue masks overlap or deformations plausibility results in an immediate rejection, while successful registrations are graded based on their best-performing aspect. The overall score $Q$ provides an interpretable, ordinal measure of RQ, evaluating deformation plausibility alongside geometric alignment metrics even in the absence of pixel-level GT. \vspace{0.2cm}
\begin{table}[t]
\scriptsize
\centering
\begin{tabular}{l | l | c | c | c }
\hline
Experts              & \multicolumn{1}{l|}{Eval. Model} & \multicolumn{1}{l|}{$w$AP} & \multicolumn{1}{l|}{$w$AR} & \multicolumn{1}{l}{$w$F1} \\ \hline
\multirow{3}{*}{E-1} & MRQA                      &        0.84                &         0.82               &    0.83                    \\
                   & MRQA-5K                      &        0.85               &      0.84                  &         0.85               \\ 
                   & DRQA                      &        0.84                &      0.85                  &         0.83               \\ 
                   & \textit{URQA (Proposed)}                    &             \textit{ 0.87  }       &  \textit{ 0.82 }                    &        \textit{  0.83  }            \\ 
   \hline
\multirow{3}{*}{E-2} & MRQA                      &      0.84                  &           0.82             &    0.83                    \\
                   & MRQA-5K                      &         0.85               &             0.84         &         0.85               \\ 
                   & DRQA                      &        0.76               &      0.79                  &         0.76               \\ 
                   & \textit{URQA (Proposed)  }                 &   \textit{0.81 }                    &      \textit{ 0.76  }               &            \textit{ 0.77    }       \\ 
   \hline
\end{tabular}
\vspace{-0.04cm}
\caption{\small Comparison of RQA accuracy using utilizing $w$: weighted metrics; AP: average precision; AR: average recall; F1: F1-score across modalities, compared against experts E-1 and E-2. \normalsize }
\label{tab1}
\end{table}
\begin{table}[tb]
\begin{minipage}{.45\linewidth}     
      \centering
      \scriptsize %
	\begin{tabular}{clc}	
	\hline
	\multicolumn{1}{l|}{Input} & \multicolumn{1}{l|}{\begin{tabular}[c]{@{}l@{}}$\Delta$ RSS\\ (MB)\end{tabular}} & \multicolumn{1}{l}{\begin{tabular}[c]{@{}l@{}}Time\\ (s)\end{tabular}} \\ \hline
	 \multicolumn{1}{l|}{512*512}                      &       \multicolumn{1}{l|}{154.16}                                &    1.86  \\
 	\multicolumn{1}{l|}{5K*5K}                      &              \multicolumn{1}{l|}{428.42}                         &    6.76                                         \\
 	 \multicolumn{1}{l|}{15K*15K}                      &                     \multicolumn{1}{l|}{2150.9}                  &           430.27                                         \\
	 \multicolumn{1}{l|}{25K*25K}                      &                \multicolumn{1}{l|}{5485.98}                       &           3420.97                  \\ \hline                   
	\end{tabular}%
	\vspace{-0.05cm}
	\caption{\small{MRQA module efficiency on a WSI pair at various sizes; default: 512.} }
	\label{tab2}
\end{minipage}%
% Add vertical space here
\hspace{0.3cm}
\begin{minipage}{.54\linewidth}
\centering
\scriptsize %
\begin{tabular}{cl}
	\hline
	\multicolumn{1}{l|}{Eval. Methods} &  \multicolumn{1}{l}{Avg. Time/Slide}  \\ \hline
	\multicolumn{1}{l|}{Experts}        & 213.64                                                                               \\
	\multicolumn{1}{l|}{MRQA}        & 4.3                                                                                 \\
	\multicolumn{1}{l|}{MRQA-5K}        & 33.48                                                                                 \\
	\multicolumn{1}{l|}{DRQA}                              & 6.3                                                                                  \\ 

	\multicolumn{1}{l|}{URQA (Proposed)    }               & 10.75                                                                              \\ \hline
	\end{tabular}%
	\vspace{-0.05cm}
	\caption{\small{Comparison of average processing time per slide (seconds) across evaluation methods. }}
	\label{tab3}
	\vspace{-0.1cm}
\end{minipage} 
\end{table}
\vspace{-0.00cm}
\vspace{-0.5cm}
\begin{figure}[]
\begin{minipage}[b]{0.4\linewidth}
  \centering
  \centerline{\includegraphics[clip, trim=0.0 4.6cm 14.0cm 0.00cm, width=3.5cm]{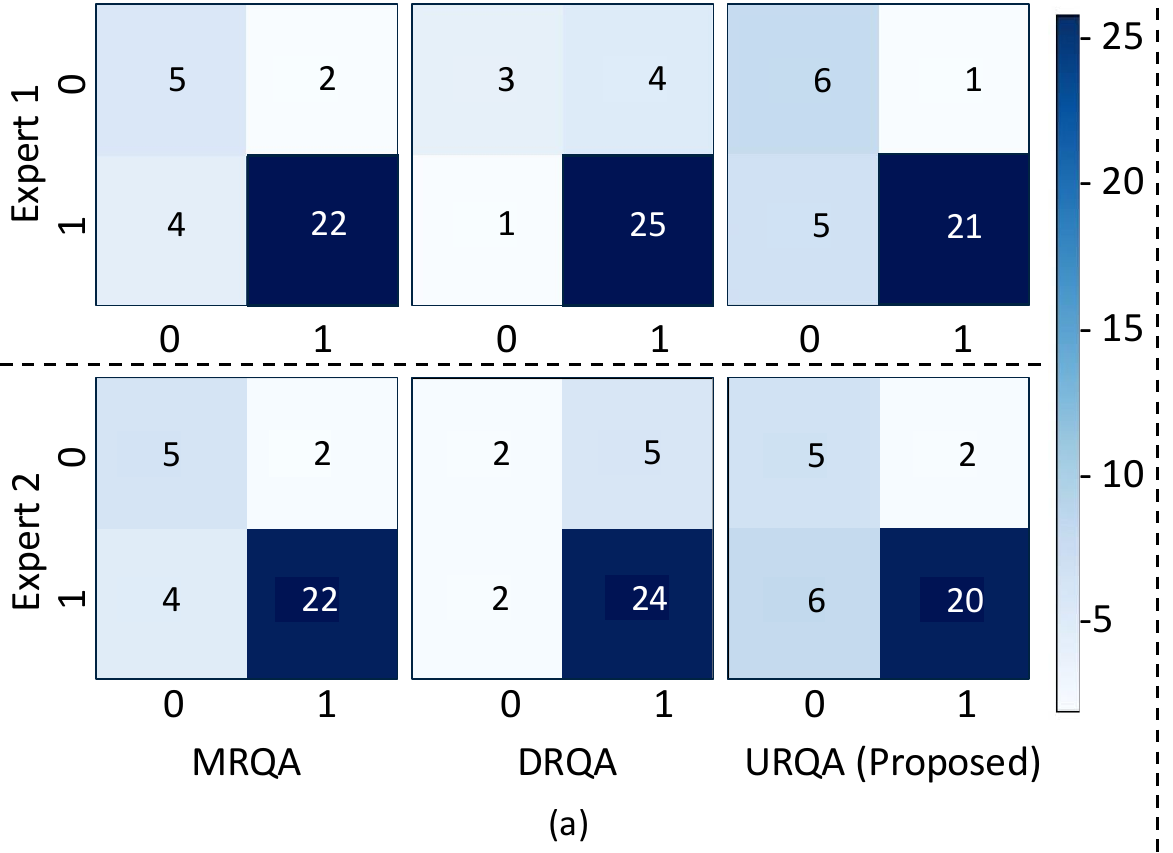}}
 %\medskip
\end{minipage}
\begin{minipage}[b]{0.6\linewidth}
  \centering
  \centerline{\includegraphics[clip, trim=0.0 4.6cm 5.7cm 0.00cm, width=5.0cm]{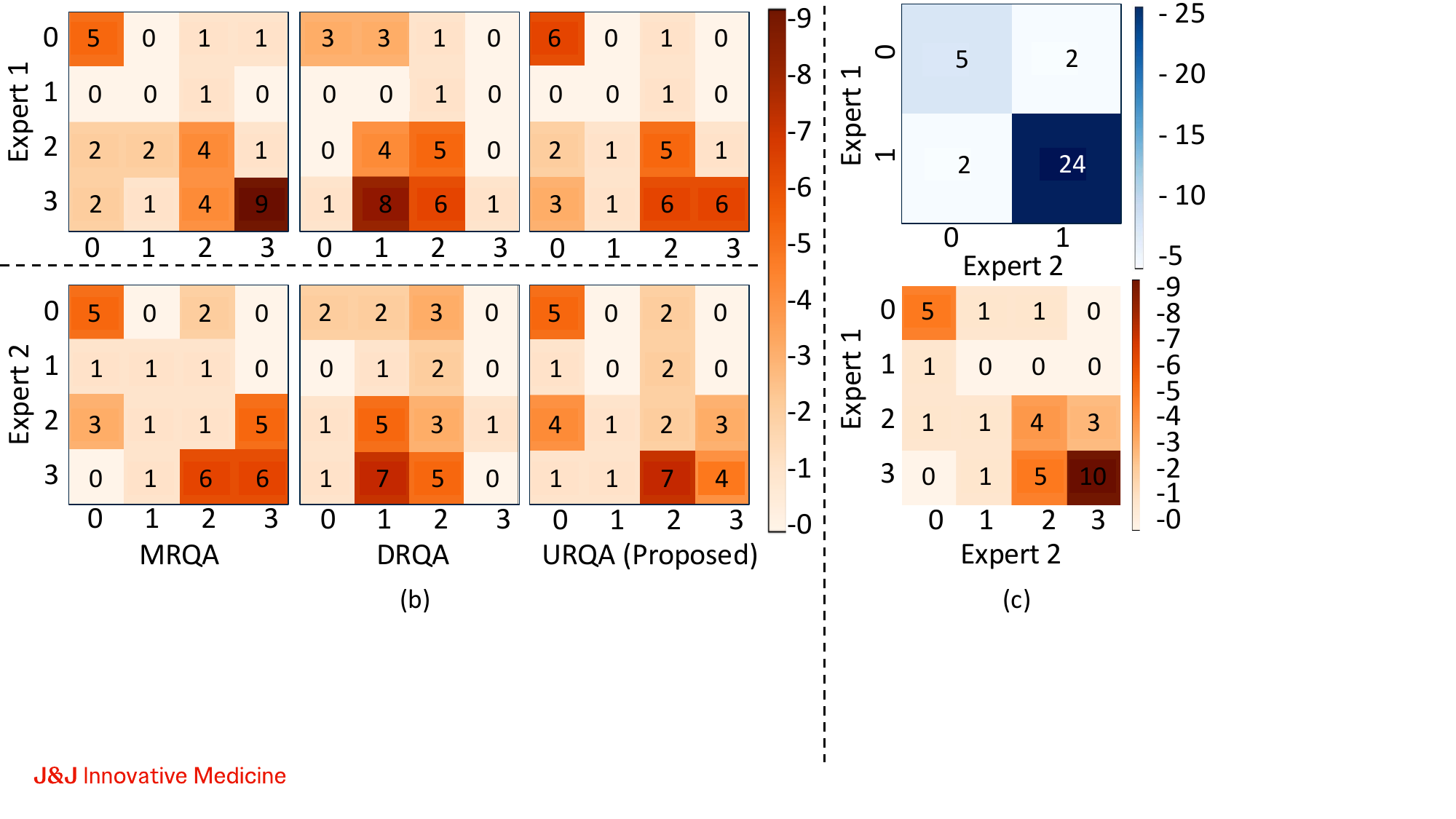}}
%  \vspace{1.5cm}
\end{minipage}
\vspace{-0.6cm}
\caption{\small Confusion matrices: (a) Binary RQA (0:Fail, Pass:1); (b) Cardinal RQA (0:Poor, 1:Fair, 2:Good, 3:Excellent); (c) Comparison of binary and cardinal assessments between experts. \normalsize}
\label{fig4}
\vspace{-0.6cm}
\end{figure}\\
\vspace{-0.1cm}
\begin{figure*}[tb]
\begin{minipage}[b]{0.45\linewidth}
  \centering
  \centerline{\includegraphics[clip, trim=0.0 2.8cm 4.85cm 0.00cm, width=8.1cm]{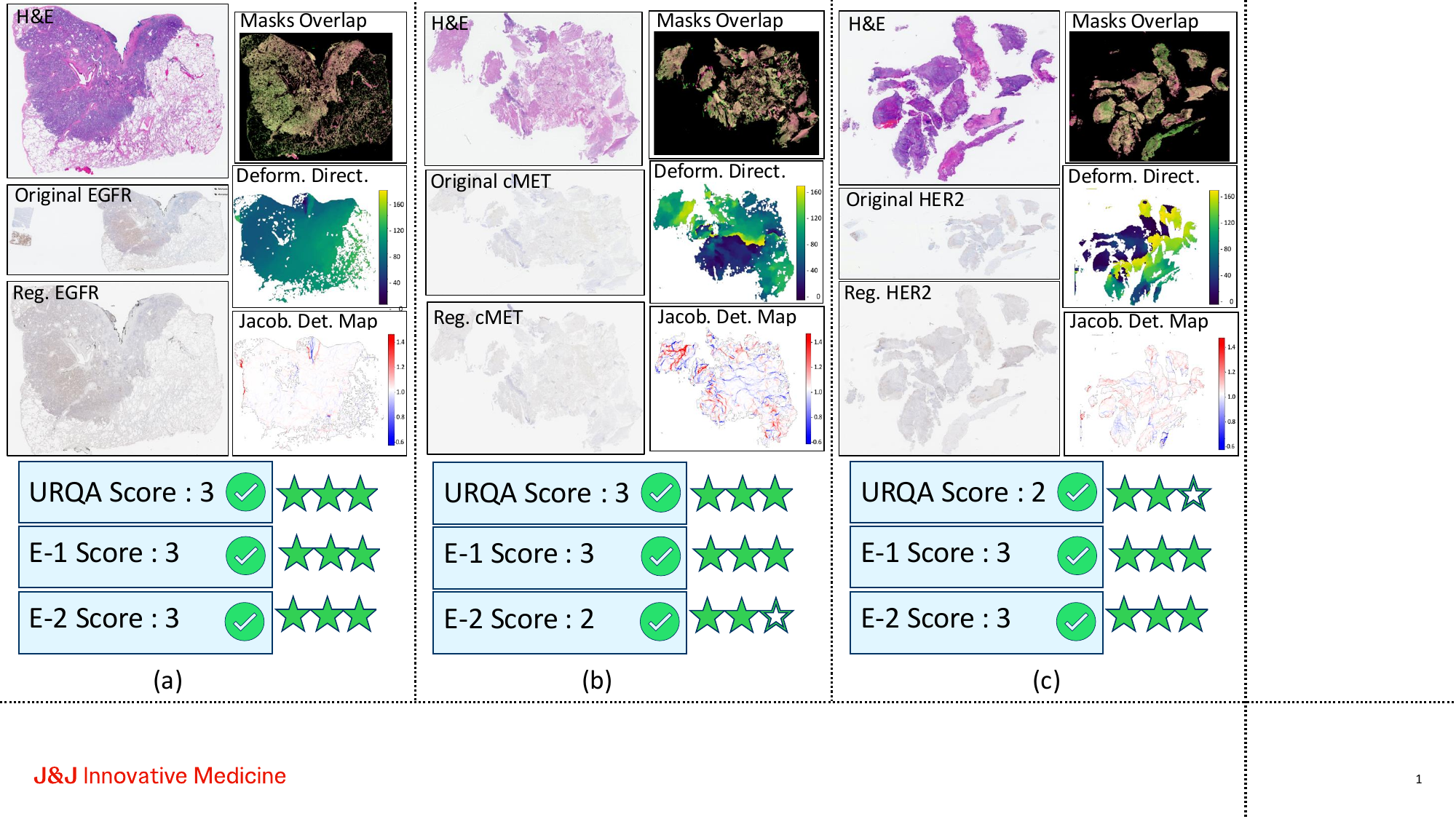}}\end{minipage}
\begin{minipage}[b]{0.52\linewidth}
  \centering
  \centerline{\includegraphics[clip, trim=0.0 2.7cm 0.2cm 0.00cm, width=9.3cm]{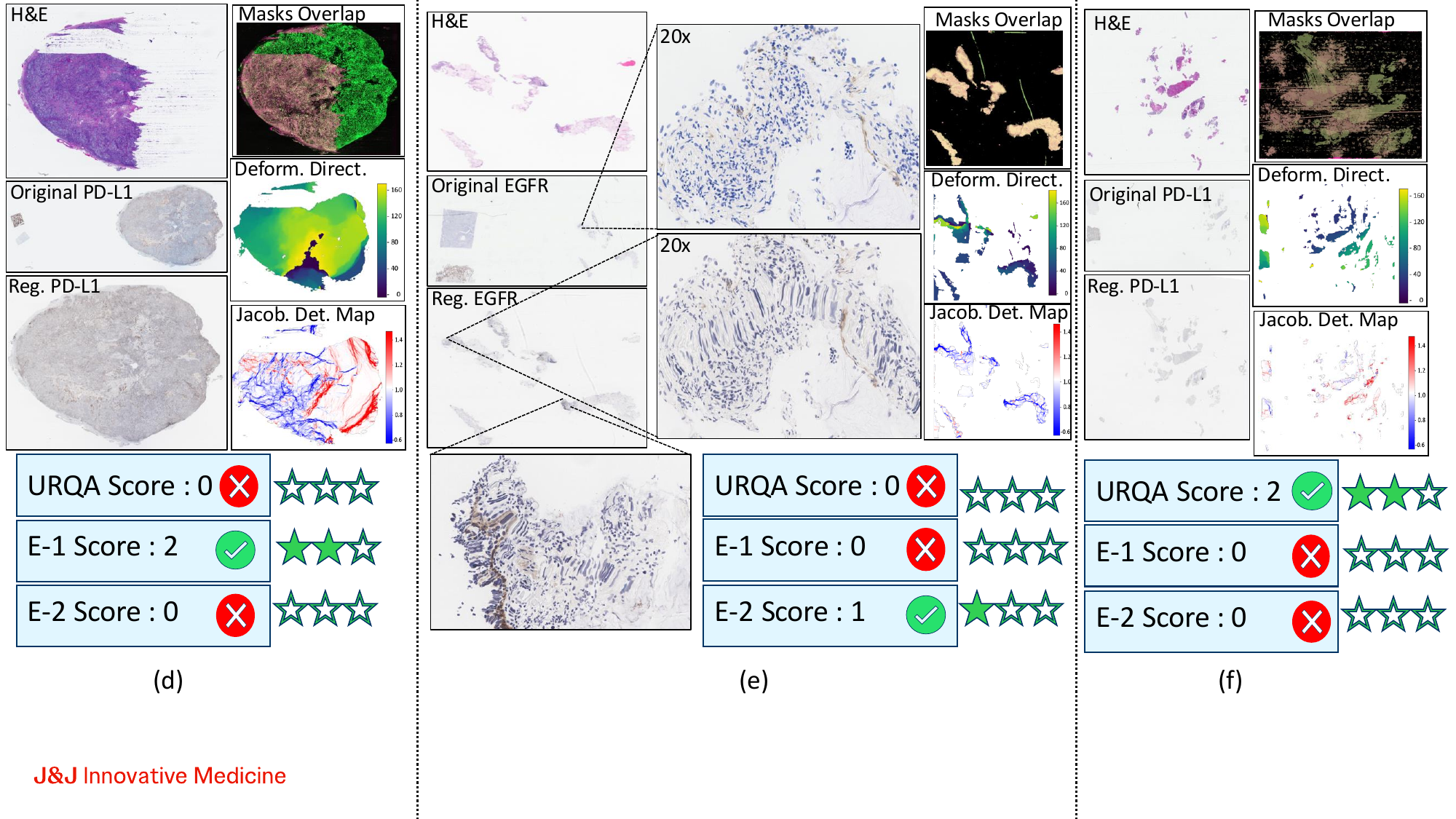}}\end{minipage}
\vspace{-0.3cm}
\caption{\small Qualitative RQA examples: (a–c) align with both experts; (d–e) show partial agreement; (f) no agreement with experts. Each displays tissue masks, deformation directions, and Jacobian maps; (e) highlights poor deformations at 20x.\normalsize}
\label{fig3}
\vspace{-0.28cm}
\end{figure*}
\section{Experimentations and Results}\label{exp}\vspace{-0.07cm}
\subsection{Dataset and Preprocessing}\label{data} \vspace{-0.07cm}
Experiments were conducted on an in-house dataset comprising 300 H\&E and IHC-stained WSI pairs, each pair from the same tissue block, covering multiple markers including 18 HER2, 90 cMET, 10 EGFR, and 182 PD-L1. About 95\% of the data consisted of sections cut between 5--10 microns apart and imaged simultaneously, while 5\% had more than 50 microns of distance between H\&E and IHC. Based on the available IHC markers ratios, 8 HER2, 10 cMET, 5 EGFR, and 10 PD-L1 slides were randomly selected for expert validation, excluding 25 slides reserved for empirical studies used to determine thresholds in Sections \ref{maskthresh}, \ref{deformthresh}. The slides were registered using VALIS's default parameters \cite{valis}.
\subsection{Results and Discussions}\label{res}
In the absence of GT, URQA's scoring was validated via multi-expert analysis and qualitative assessments as below.\\
\textbf{Experts Validation and Comparative Analysis:}
URQA was evaluated on 300 H\&E–IHC registered slide pairs, of which approximately 40\% passed the automated RQA. To verify this, a subset of 33 slides was independently reviewed by two experts, who were provided with the registered H\&E–IHC pairs and the corresponding original IHC WSIs. The two experts—E-1, a board-certified US-AP/CP pathologist, and E-2, an ASCP-licensed histotechnologist—graded each pair into a binary \textit{Pass}(1) or \textit{Fail}(0), and into four categories: (0:\textit{Poor}, 1:\textit{Fair}, 2:\textit{Good}, 3:\textit{Excellent}) based on visual inspection of glandular and structural alignment. Fig.\ref{fig4}(a) illustrates the binary confusion matrices comparing expert evaluations with the proposed model and its two modules: MRQA and DRQA. All methods showed good agreement with expert ratings, although URQA achieved the lowest false positives and the highest true negatives with E-1, while maintaining comparable true positives. The MRQA module aligned slightly better with E-2; however, inter-observer variability is also evident within expert ratings, as shown in Fig.\ref{fig4}(c). We also compared the overall categorical RQ across all modalities, as shown in Fig.\ref{fig4}(b). The subjective nature of classifying RQ is evident; nonetheless, URQA has attempted to match the RQ provided by experts to some extent.\\
Table \ref{tab1} shows that integrating geometric and deformation metrics (URQA) improves weighted AP over individual modules for E-1. While DRQA alone struggles due to limited spatial context, its combination with masks-based metrics results in a more reliable and interpretable RQA across modalities. Although MRQA aligns better with E-2, qualitative analysis highlights the importance of combining it with DRQA for more robust RQA. Additionally, MRQA-5K is the same as MRQA but evaluates RQ at 5K resolution.\\
\textbf{Qualitative Analysis:}
Fig.\ref{fig3} shows examples: (a–c), where assessments align with both experts; (d–e), with partial agreement; and (f), a failure diverging from both. Visualizations include tissue-mask overlaps, deformation directions, and Jacobian maps to interpret URQA scores. In (a–c), high overlap, coherent deformation directions, and stable Jacobians indicate well-registered, plausible transformations. In (d), while E-1 rated it as \textit{Pass}, the model was able to identify poor registration due to tissue damage, despite smooth deformations. In (e), URQA and E-1 scored it as \textit{Poor} because of localized excessive deformations, shown at 20x in the image, while E-2 rated it as \textit{Pass}. Finally, (f) shows a failure caused by imperfect mask generation, highlighting the importance of accurate mask extraction for reliable assessment.\\
\textbf{Computational Efficiency:}
URQA’s efficiency was assessed by measuring the runtime and memory use (resident set size, RSS) of its MRQA module on a single WSI (73,728×49,152; see Table~\ref{tab2}). Runtime and RSS were recorded before and after processing. Using PyVIPS \cite{valis} for memory-efficient thumbnail loading, slides were processed at resolutions of 512, 5K, 15K, and 25K pixels. Although higher resolutions increase runtime and memory consumption, MRQA at 512 px provides fast, resource-efficient evaluation that significantly reduces resource demands while maintaining comparable accuracy (Table~\ref{tab1}, MRQA = MRQA-512), making it suitable for large-scale deployment. Furthermore, Table~\ref{tab3} compares the average evaluation times across all slides provided to experts, showing that automated RQA methods drastically lower processing time compared to expert assessments, facilitating scalable analysis of large histopathology datasets.\\ \vspace{-0.54cm}
\section{Conclusion and Future Work}
\label{conclusion} \vspace{-0.1 cm}
This work introduces URQA, a novel, fast, scalable, and fully automated framework for assessing registration quality between H\&E and IHC WSIs, without GT. It combines tissue masks and compressed deformation fields to evaluate geometric and physical plausibility. Validated on VALIS-registered slides, future directions include improving tissue-mask segmentation, extending to other registration models like DeepHistReg \cite{deephistreg} and ANTs \cite{ants}, and testing on public datasets such as ANHIR \cite{anhir} and ACROBAT \cite{ACROBAT} for broader generalization. Additionally, URQA will be explored for downstream tasks like cell-type mapping, multiplex staining, and patch-level feature learning, supporting large-scale histology analysis and clinical decision-making.

\section{Acknowledgments}
\label{sec:acknowledgments}
This work was sponsored by Johnson \& Johnson. Thanks to Michael Sharp from Johnson \& Johnson Oncology Translational Research group for providing one of the expert evaluations in this work.
\section{Compliance with Ethical Standards}
\label{sec:ethic}
The patient imaging data used in this study were de-identified before use, as required by law.
\section{Disclosure of Interest}
\label{sec:disclosure}
The authors have no competing interests to declare that are relevant to the content of this article.

\bibliographystyle{IEEEbib}
\bibliography{draft_ref}

@Article{ants,
  author= "Tustison, Nicholas J. and Cook, Philip A. and Holbrook, Andrew J. and Johnson, Hans J. and Muschelli et al., John",
  journal= "Scientific Reports", 
  title= "The ANTsX ecosystem for quantitative biological and medical imaging", 
  year= "2021",
  volume= "11",
  number= "1",
  pages= "9068"
 }

@Article{voxelmorph,
  author= "Guha Balakrishnan and Amy Zhao and Mert R. Sabuncu and John Guttag and Adrian V. Dalca",
  journal= "IEEE TMI: Transactions on Medical Imaging", 
  title= "VoxelMorph: A Learning Framework for Deformable Medical Image Registration", 
  year= "2019"
 }

@Article{deephistreg,
  author= "Marek Wodzinski and Henning Müller",
  journal= "Computer Methods and Programs in Biomedicine", 
  title= "DeepHistReg: Unsupervised Deep Learning Registration Framework for Differently Stained Histology Samples", 
  volume = "198",
pages = "105799",
year = "2021"
 }

@Article{anhir,
  author= "Borovec, Jiří and Kybic, Jan and Arganda-Carreras, Ignacio and Sorokin, Dmitry V. and Bueno et al., Gloria",
  journal= "IEEE Transactions on Medical Imaging", 
  title= "ANHIR: Automatic Non-Rigid Histological Image Registration Challenge", 
  year= "2020",
  volume= "39",
  number= "10",
  pages= "3042-3052"
 }

@Article{elastix,
  author= "Klein, Stefan and Staring, Marius and Murphy, Keelin and Viergever, Max A and Pluim, Josien P W",
  journal= "IEEE Transactions on Medical Imaging", 
  title= "elastix: a toolbox for intensity-based medical image registration", 
  year= "2010",
  volume= "29",
  number= "1",
  pages= "196-205"
  }

@Article{valis,
  author= "Gatenbee, Chandler D. and Baker, Ann-Marie and Prabhakaran, Sandhya and Swinyard, Ottilie and Slebos et al., Robbert J. C.",  
  journal= "Nature Communications", 
  title= "Virtual alignment of pathology image series for multi-gigapixel whole slide images", 
  year= "2023",
  volume= "14",
  number= "1"
 }

@Article{deform,
  author= "Avants, BB and Epstein, CL and Grossman, M and Gee, JC",  
  journal= "Medical Image Analysis", 
  title= "Symmetric diffeomorphic image registration with cross-correlation: evaluating automated labeling of elderly and neurodegenerative brain", 
  year= "2008",
  volume= "12",
  number= "1",
  pages= "26-41"
   }

@Article{eval1,
author = "Cagni, Elisabetta and Botti, Andrea and Orlandi, Matteo and Galaverni, Marco and Iotti et al., Cinzia",
title = "Evaluating the Quality of Patient-Specific Deformable Image Registration in Adaptive Radiotherapy Using a Digitally Enhanced Head and Neck Phantom",
journal = "Applied Sciences",
volume = "12",
year = "2022",
number = "19"
}

@Article{GeoMetrics2Path,
author = {Md. Ziaul Hoque and Anja Keskinarkaus and Pia Nyberg and Taneli Mattila and Tapio Seppänen},
title = "Whole slide image registration via multi-stained feature matching",
journal = "Computers in Biology and Medicine",
year = "2022",
volume = "144",
pages = "105301"
}

@InProceedings{vims,
author= "Shikha Dubey and Yosep Chong and Beatrice Knudsen and Shireen Y Elhabian",
title= "Vims: Virtual immunohistochemistry multiplex staining via text-to-stain diffusion trained on uniplex stains.",
booktitle= "International Conference on Medical Image Computing and Computer Assisted Intervention (MICCAI) Workshop",
year= "2024"
}

@Article{PATIL2023,
author = "Patil, Abhijeet and Diwakar, Harsh and Sawant, Jay and Kurian, Nikhil Cherian and Yadav et al., Subhash",
title = "Efficient quality control of whole slide pathology images with human-in-the-loop training",
journal = "Journal of Pathology Informatics",
volume = "14",
pages = "100306",
year = "2023"
}

@Article{ACROBAT,
title = "The ACROBAT 2022 challenge: Automatic registration of breast cancer tissue",
journal = "Medical Image Analysis",
volume = "97",
pages = "103257",
year = "2024",
author = "Philippe Weitz and Masi Valkonen and Leslie Solorzano and Circe Carr and Kimmo Kartasalo et al."
}

@InProceedings{GeoMetrics,
title = {HISTOPATHOLOGY IMAGE REGISTRATION BY INTEGRATED TEXTURE AND SPATIAL PROXIMITY BASED LANDMARK SELECTION AND MODIFICATION},
booktitle = {IEEE International Symposium on Biomedical Imaging (ISBI)},
pages = {1827-1830},
year = {2021},
author = {Pangpang Liu and Fusheng Wang and George Teodoro and Jun Kong}
}

@Article{GeoMetrics3,
title = {A high-precision hierarchical registration approach for stain- and scanner-independent colocalization on whole slide images in histopathology},
journal = {Health Information Science and Systems},
volume = {13},
year = {2025},
author = {Tom Bisson and Michael Franz and Tim-Rasmus Kiehl and Peter Boor and Peter Hufnagl and Norman Zerbe }
}

@Article{ASYLUM,
title = {A synthetic lung model (ASYLUM) for validation of functional lung imaging methods shows significant differences between signal-based and deformation-field-based ventilation measurements},
journal = {Frontiers in Medicine},
volume = {11},
year = {2024},
author = {Andreas Voskrebenzev and Marcel Gutberlet and Filip Klime et al. }
}

@ARTICLE{otsu,
  author={Otsu, Nobuyuki},
  journal={IEEE Transactions on Systems, Man, and Cybernetics}, 
  title={A Threshold Selection Method from Gray-Level Histograms}, 
  year={1979},
  volume={9},
  number={1},
  pages={62-66}}

\end{document}